\documentclass[10pt,twocolumn,letterpaper]{article}

\usepackage[pagenumbers]{cvpr} 

\usepackage{graphicx}
\usepackage{amsmath}
\usepackage{amssymb}
\usepackage{booktabs}
\usepackage{caption}
\usepackage{multirow}

\usepackage{array}
\makeatletter
\newcommand{\thickhline}{%
    \noalign {\ifnum 0=`}\fi \hrule height 0.8pt
    \futurelet \reserved@a \@xhline
}

\usepackage[pagebackref,breaklinks,colorlinks]{hyperref}

\usepackage[capitalize]{cleveref}
\crefname{section}{Sec.}{Secs.}
\Crefname{section}{Section}{Sections}
\Crefname{table}{Table}{Tables}
\crefname{table}{Tab.}{Tabs.}

\def\cvprPaperID{*****} 
\def\confName{CVPR}
\def\confYear{2024}

\begin{document}


\title{UniBrain: Unify Image Reconstruction and Captioning All in One Diffusion Model from Human Brain Activity}

\author{Weijian Mai\textsuperscript{1},  Zhijun Zhang\textsuperscript{1*}\\
\textsuperscript{1}South China University of Technology \\
{\tt\small drzhangzhijun@gmail.com}
\thanks{Corresponding author}
}
\maketitle

\begin{abstract}
   Image reconstruction and captioning from brain activity evoked by visual stimuli allow researchers to further understand the connection between the human brain and the visual perception system. While deep generative models have recently been employed in this field, reconstructing realistic captions and images with both low-level details and high semantic fidelity is still a challenging problem.  In this work, we propose \textit{\textbf{UniBrain:} \textbf{Uni}fy Image Reconstruction and Captioning All in One Diffusion Model from Human \textbf{Brain} Activity}. For the first time, we unify image reconstruction and captioning from visual-evoked functional magnetic resonance imaging (fMRI) through a latent diffusion model termed Versatile Diffusion. Specifically, we transform fMRI voxels into text and image latent for low-level information and guide the backward diffusion process through fMRI-based image and text conditions derived from CLIP to generate realistic captions and images.  UniBrain outperforms current methods both qualitatively and quantitatively in terms of image reconstruction and reports image captioning results for the first time on the Natural Scenes Dataset (NSD) dataset. Moreover, the ablation experiments and functional region-of-interest (ROI) analysis further exhibit the superiority of UniBrain and provide comprehensive insight for visual-evoked brain decoding.

\end{abstract}


\section{Introduction}
\label{sec:intro}

How does the human brain process and understand external visual stimuli? Image reconstruction and captioning from brain activity (i.e. functional magnetic resonance imaging (fMRI)) may give us a plausible answer. Image reconstruction from visual-evoked fMRI could potentially be used to create brain-computer interfaces that can decode what someone is seeing or even help individuals with certain neurological conditions communicate their visual experiences. Instead of reconstructing the actual visual image, image captioning from visual-evoked fMRI focuses on generating descriptive captions or textual explanations of the visual stimuli, which could have applications in helping individuals with communication difficulties express their visual experiences or in developing brain-computer interfaces that can translate brain activity into natural language descriptions. Combining image reconstruction and captioning from visual-evoked fMRI would create a powerful system that not only reconstructs the visual image that a participant is viewing but also generates a descriptive caption or textual explanation of that image. This combined approach could provide a more comprehensive understanding of the participant's visual experience, bridging the gap between neural activity and language comprehension. However, to the best of our knowledge, no previous work has attempted to design a unified framework for deeper and more comprehensive analysis.


Image reconstruction from visual-evoked fMRI is an intriguing yet difficult task due to the limited understanding of potential neural representations and the typically small sample size of neural data. \cite{nature-id}\cite{neuron-visual}\cite{neuron-bayesian}.
Researchers have recently begun to focus on image reconstruction tasks by utilizing deep generative models (i.e. generative adversarial networks (GANs))\cite{Mind-Reader}\cite{shape-semantic}\cite{gan2018}\cite{bigbigan}\cite{IC-GAN}\cite{vg-gan}, as well as self-supervised learning \cite{ssfMRI-1}\cite{ssfMRI-2}.
Furthermore, recent research has improved the semantic fidelity by explicitly incorporating semantic content as supplementary information for reconstruction \cite{shape-semantic}\cite{Mind-Reader}. Nevertheless, these investigations necessitate training from scratch or fine-tuning complex deep generative models using limited neural data. This imbalance between data and parameters limits model performance in terms of pixel and semantic fidelity. Image captioning needs to build connections between multiple modalities of brain data, stimulus images, and semantic text. Contextual associations of text data often need to be captured using sequence models. Hence, current researchers tend to use a hybrid structure of Convolutional Neural Networks (CNN) and sequence models (i.e. Recurrent Neural Networks(RNN), Transformer) to decode visual-evoked fMRI into language step by step \cite{PT-LDM}\cite{CNN-Transformer}.

In recent years, diffusion models (DMs) \cite{DDPM}\cite{DM2}\cite{DM3} have garnered significant interest as deep generative models. DMs have demonstrated exceptional performance in various tasks, including image colorization \cite{img_color}, image super-resolution  \cite{img_resol}, conditional image generation \cite{CDM1}\cite{CDM2}\cite{CDM3}, and other relevant tasks\cite{DM-Speech}\cite{DM-img-trans}, establishing themselves as the leading models in these areas. As an improvement, latent diffusion models (LDMs) \cite{ldm} have achieved computational expense reduction through the utilization of the latent space produced by their autoencoding elements. This allows for more efficient calculations during training and inference, as well as the capability to generate high-resolution images with high semantic fidelity. On the one hand, Takagi \textit{et al.} \cite{Takagi}  first attempted to map fMRI to the input space of a text-guided latent diffusion model (Stable Diffusion) and generate high-fidelity images without training or fine-tuning deep neural networks. However, the reconstructed images of this model are still not semantic and natural enough, probably due to the limited single-modality (only text) condition. On the other hand, to the best of our knowledge, no study has used diffusion models for image captioning from visual-evoked fMRI, let alone performing fMRI-based image reconstruction and captioning all in one diffusion model.


To tackle these issues,  we propose a multi-task diffusion framework with multi-modality conditions (text and image) termed UniBrain based on Versatile Diffusion (VD) \cite{VD} for visual-evoked fMRI-based image reconstruction and captioning. UniBrain is capable of generating images and descriptive captions from human brain activity (fMRI) evoked by visual stimuli (Figure \ref{fig_1}). The contributions of this paper are as follows:

\begin{itemize}
    \item By mapping fMRI to the input space of diffusion, UniBrain is capable of generating high-quality images and captions from fMRI without any training and fine-tuning of the deep learning model.
    \item UniBrain uses multi-modality conditions (image and text) to guide the generation of images and captions, significantly improving semantic fidelity compared to typical text-guide diffusion models.
    \item  For the first time, a unified model UniBrain is proposed for visual-evoked fMRI-based image reconstruction and captioning. While achieving state-of-the-art on each task, UniBrain also contributes to a more comprehensive analysis of how human brain processes and understands external visual stimuli.
\end{itemize}

\begin{figure*}[!t]
\centering
\includegraphics[width=\linewidth]{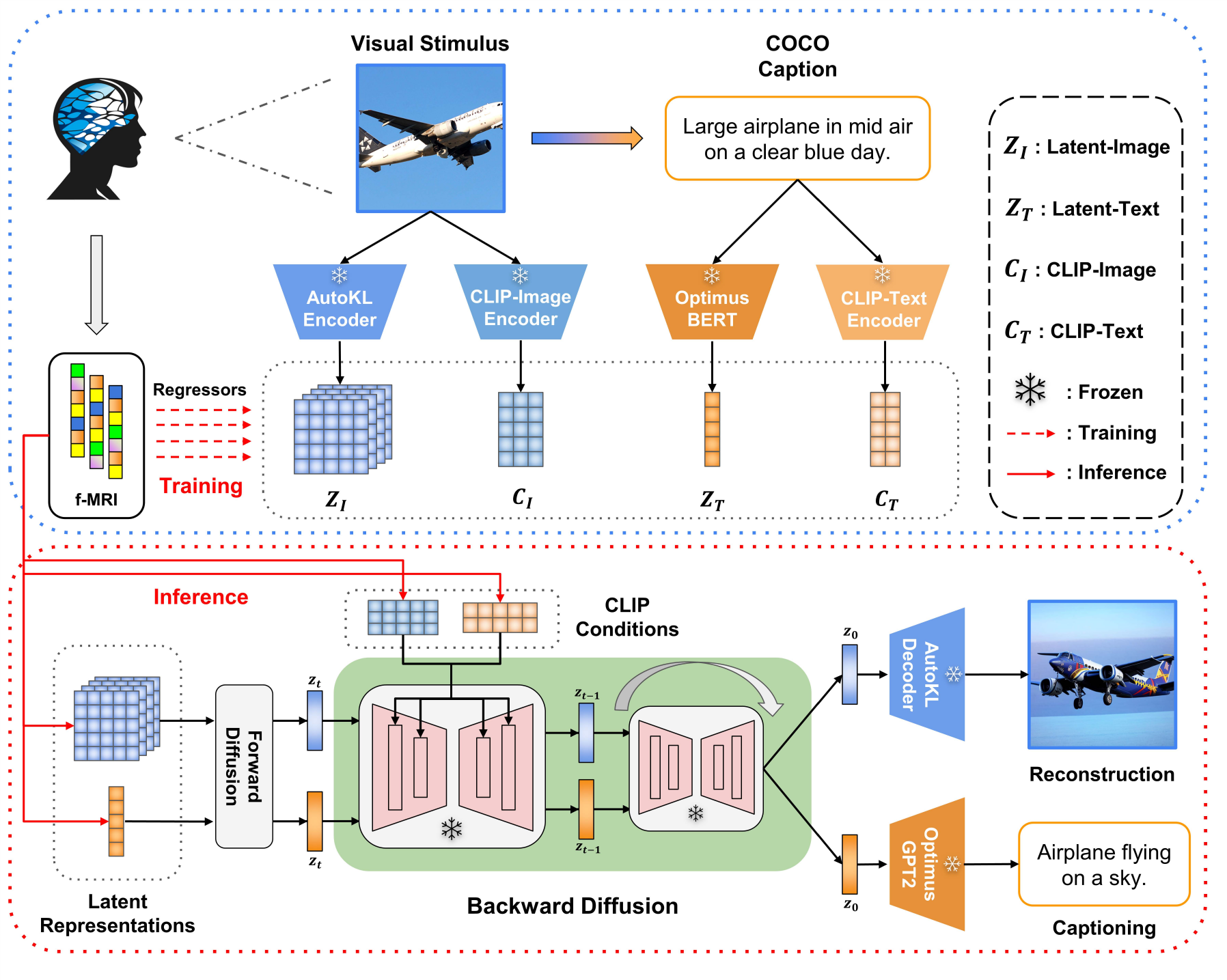}
\caption{Framework of UniBrain. In the training stage (top), we map fMRI voxels to Latent-Image $Z_I$, Latent-Text $Z_T$, CLIP-Image $C_I$, and CLIP-Text $C_T$ via separated regressors. In the inference stage (bottom), we infer low-level latent ($Z_I$ and  $Z_T$) and high-level CLIP conditions ($C_I$ and $C_T$) from test fMRI data, which are fed into the pretrained versatile diffusion model for image reconstruction and captioning.}
\label{fig_2}
\end{figure*}

\section{Methodology}
\label{sec:method}

\subsection{Latent Diffusion Models}
\label{subsec: ldm}

Diffusion Models (DMs) refer to probabilistic generative models that utilize iterative denoising to recover a sampled variable from Gaussian noise and transform it into a sample conforming to the learned data distribution. The gradual incorporation of Gaussian noise through the forward diffusion process disrupts the inherent structure of the raw training data. The noisy variant of the initial input $x_0$ at each time point is defined as $x_t = \sqrt{\alpha_t}x_0 + \sqrt{1-\alpha_t}\epsilon_t$. The reverse diffusion process is learned via a neural network (Denoising U-Net) to predict and remove noise from the noisy input so as to retrieve the original variables. This is done by minimizing the loss function as follows:
\begin{eqnarray}\label{eqn.dm_loss}
L_{DM}=E_{x_0, \epsilon\sim\mathcal N(0,1) ,t}\left [||\epsilon-\epsilon_\theta(x_t,t)||^2\right]
\end{eqnarray}
where $\alpha$ represents the control factor for adding noise, $\epsilon$ is the true Gaussian noise, $\epsilon_\theta(\cdot)$ represents the neural network trained to predict the noise, and $t\in\{1,2,...,T\}$ denotes the time step.

However, DMs that operate in pixel space are computationally expensive. LDMs overcome this limitation by compressing the input using an autoencoder $E(\cdot)$, which is trained on a large-scale image dataset to learn a compressed latent representation $z_0$ from image $x_0$ ($z_0 = E(x_0)$). In doing so, the forward diffusion process can be denoted as $z_t = \sqrt{\alpha_t}z_0 + \sqrt{1-\alpha_t}\epsilon_t$. The reverse diffusion process of LDM is similar to that of DM except that it performs in a latent space with additional conditions. This objective function is defined as below:
\begin{eqnarray}\label{eqn.ldm_loss}
L_{LDM}=E_{z_0, c, \epsilon\sim\mathcal N(0,1) ,t}\left [||\epsilon-\epsilon(z_t,t,\tau_\theta(c)||^2\right]
\end{eqnarray}
where $\tau_\theta(c)$ is the conditioning input for U-Net. The crucial aspect of this procedure lies in its potential to guide the inverse diffusion process using various conditions (e.g. labels, captions, images and semantic maps). The process of conditioning is accomplished by integrating conditions $\tau_\theta(c)$ within the cross-attention block of the denoising U-Net model. The denoised latent variable derived from the reverse diffusion is passed through the pretrained decoder $D(\cdot)$ to produce a high-quality image.

Versatile Diffusion (VD) \cite{VD} is a multi-flow multimodal latent diffusion model capable of producing diverse types of results (e.g. image and text) guided by CLIP features derived from images, text, or image-text pairs. Moreover, VD offers an alternative approach that involves commencing the reverse diffusion process with latent variables acquired from a specific text and image, as opposed to using entirely random distribution-based initialization. The Versatile Diffusion model utilized in our research was trained on two extensive datasets: Laion2B-en \cite{Laion} and COYO-700M \cite{Coyo}, both of which consist of a substantial number of image-text pairs. The CLIP network employed in VD utilizes the transformer framework (ViT-L/14) and undergoes pretraining via an extensive contrastive task \cite{Clip}.

\subsection{Image Reconstruction and Captioning}
\label{subsec: recon}

Based on versatile diffusion, we propose a multi-task multi-modality model termed UniBrain for image reconstruction and captioning from visual-evoked fMRI.  The framework of UniBrain is shown in Figure \ref{fig_2}. Specifically, following the text-to-text and image-to-image pipelines, we perform a pretrained AutoKL encoder and Optimus BERT \cite{bert} encoder on visual stimuli and corresponding COCO \cite{COCO} captions (random selected from five text annotations) to derive low-level latent representations of image and text (represent as $Z_I$ and $Z_T$), respectively. Furthermore, the pretrained image and text encoders from CLIP are utilized to derive high-level image and text conditions (represent as $C_I$ and $C_T$) from visual stimuli and COCO captions, respectively. 

In the training state (Figure.\ref{fig_2} (top)), we train regressors between i) fMRI and Latent-Image $Z_I$ (4x64x64); ii) fMRI and Latent-Text $Z_T$ (1x768); iii) fMRI and CLIP-Image $C_I$ (257x768); iv) fMRI and CLIP-Text $C_T$ (77x768). Note that all encoders are pretrained on large-scale data and frozen in this work. That is, no training or fine-tuning of complex deep neural networks is needed, we only train four tiny regression models. Each regressor maps fMRI voxels to their corresponding distributions in the training state and infers target distributions of test samples in the testing state with frozen weights (Figure.\ref{fig_2} (bottom)).

For image reconstruction, we follow the image-to-image pipeline with multi-modality CLIP conditions  (both CLIP-Text and CLIP-Image). First, we infer the low-level image latent representation $Z_I$ from test fMRI and pass through the forward diffusion process, deriving the noisy latent representation $z_t$ at time t. Next, we guide the backward diffusion process by adding the inference high-level CLIP-Text $C_T$ and CLIP-Image $C_I$ conditions in the frozen U-Net. The reconstructed image latent representation $z_0$ is derived after all diffusion steps and given as input to the pretrained AutoKL decoder to generate the high-resolution image.

For image captioning, we follow the text-to-text pipeline with multimodal CLIP conditions. Likewise, we infer the text latent representation $Z_T$ from test fMRI and guide the backward diffusion with CLIP conditions ($C_T$ and $C_I$) to reconstruct text latent representation $z_0$, which is given as input to the pretrained Optimus GPT2 \cite{gpt2} decoder to generate the descriptive caption for visual stimulus. we found that UniBrain tended to produce repeated sentences in generated captions, which is also reported in the original paper of VD. Hence, we detect and delete the repeated sentences automatically from generated captions for the final results.

Note that we perform CLIP-Image and CLIP-Text conditions to dual-guide the reverse diffusion process in every diffusion step, where the cross-attention matrices for both conditions are mixed through linear interpolation. We define a mixing rate $mix$ between CLIP-Image and CLIP-Text, which may vary from task to task. When $mix=0.6$, it means that the relative strength of CLIP-Image is 0.6 and that of CLIP-Text is 0.4.

\section{Experiments}
\label{sec:exp}

\subsection{Dataset}
\label{subsec: dataset}

We employed the publicly accessible Natural Scenes Dataset (NSD) \cite{NSD}, an extensive 7T fMRI dataset gathered from 8 subjects viewing images from the COCO dataset \cite{COCO}. Subjects viewed each image for 3 seconds while indicating whether they had encountered the image at any prior juncture during the course of the experiment. In this work, we focused on 4 subjects (Sub-1, Sub-2, Sub-3, and Sub-7) who finished all viewing trials. The training set comprised a total of 8,859 images and 24,980 fMRI trials, with the possibility of up to 3 repetitions per image. Meanwhile, the test set encompassed 982 images and 2,770 fMRI trials. In cases where images had multiple repetitions, the fMRI trials were averaged. Note that the test images remained consistent across all subjects, whereas distinct training images were adopted. For semantic content, we utilized corresponding captions that are randomly selected from five text annotations from the COCO dataset.

We utilized the preprocessed scans from NSD for functional data, with a resolution of 1.8 mm. Our analysis involved employing single-trial beta weights derived from generalized linear models, along with region-of-interest (ROI) data specific to early and higher (ventral) visual regions as supplied by NSD. The ROI comprises the following voxel counts for the respective four subjects: [15724, 14278, 13039, 12682]. To access elaborate fMRI preprocessing procedures and further information, kindly refer to the original paper \cite{NSD} as well as the NSD website\footnote{https://naturalscenesdataset.org}.

\begin{figure*}[!t]
\centering
\includegraphics[width=\linewidth]{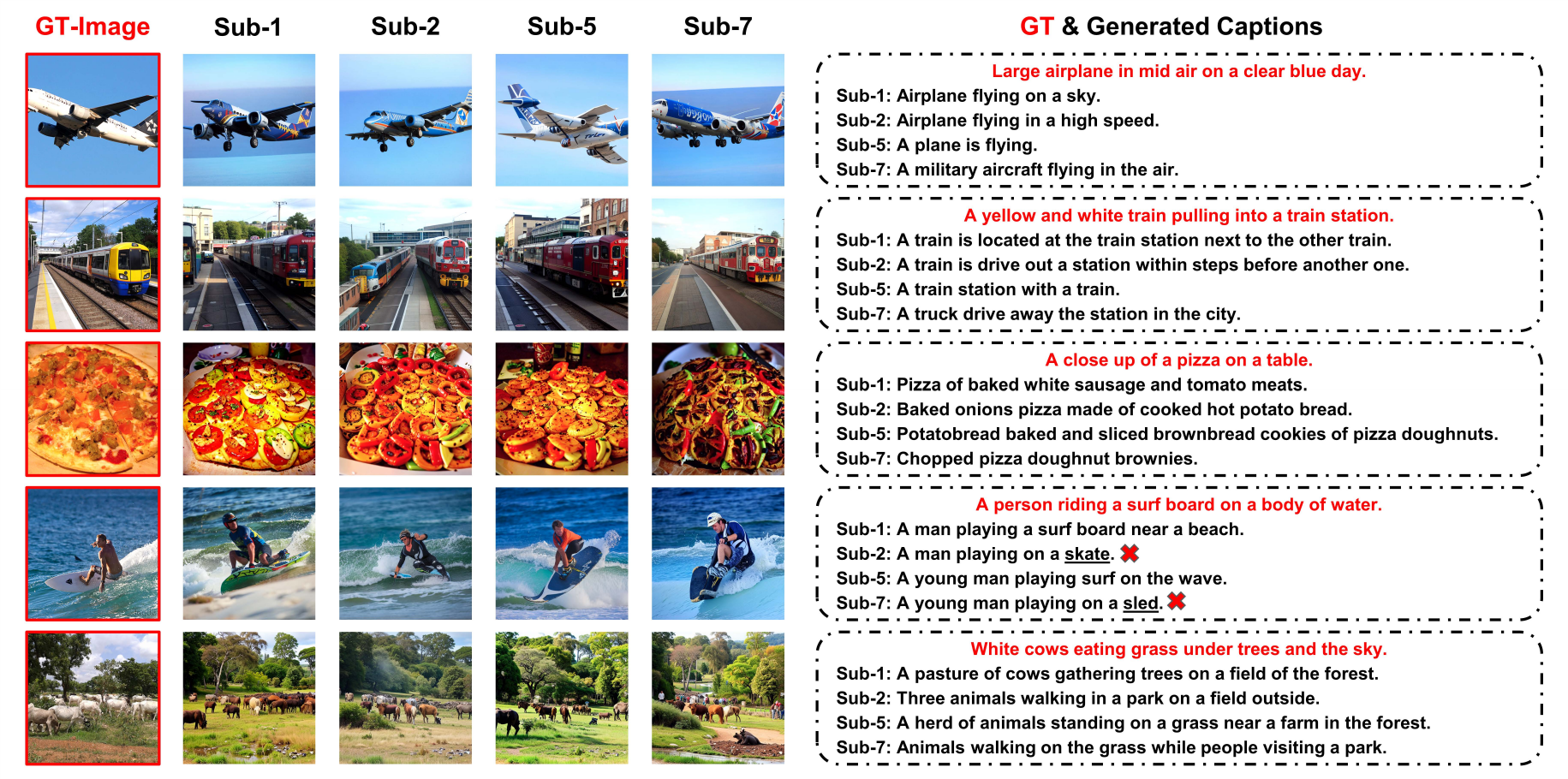}
\caption{Samples of fMRI-based image reconstruction and captioning from UniBrain. Left: Ground-True images (red box) and generated images from fMRI for all subjects. Right: Ground-True captions (red font) and generated captions from fMRI for all subjects. A red cross next to a font means a mismatch with the original caption in semantics. The presented captions are processed by automatically detecting and removing duplicate sentences from the generated captions.}
\label{fig_3}
\end{figure*}

\subsection{Implementation}
\label{subsec: imple}
UniBrain is implemented based on the versatile diffusion model\footnote{https://github.com/SHI-Labs/Versatile-Diffusion}. The regression model performed in UniBrain is Ridge regression, which is a linear regression technique that is used to handle the problem of multicollinearity (high correlation) among the predictor variables in a regression model. In the diffusion process, we use 50 diffusion steps with a diffusion strength of 0.75 for both text and image latent representations reconstruction. However, the condition mixing rate $mix$ in image reconstruction is different from that in captioning tasks. Specifically, We chose $mix=0.6$ for image reconstruction and $mix=0.9$ for image captioning based on the model performance.

\subsection{Evaluation Metric}
\label{subsec: eval_metric}

To make quantitative comparisons with current models, we introduce various vision and text evaluation metrics in both low-level and high-level aspects here and report their results in the next session.

\subsubsection{Vision Metric}
\label{subsubsec: vis_metric}
 \textbf{Low-Level:} Low-level image features provide basic information about the visual content and structure of the image, measured by: i){\textbf{PixCorr:}} Pixel-level correlation of reconstructed and ground-truth images; ii){\textbf{SSIM:}} Structural similarity index \cite{SSIM}; iii){\textbf{AlexNet:}} AlexNet-2 and AlexNet-5 are the 2-way comparisons of the second (early) and fifth (middle) layers of AlexNet \cite{AlexNet}, respectively.
    
    \textbf{High-Level:} High-level image features capture semantic information, object relationships, and contextual understanding of the image, measured by: i){\textbf{Inception:}} A two-way comparison of the last pooling layer of InceptionV3 \cite{Inception}; ii){\textbf{CLIP:}} A two-way comparison of the output layer of the CLIP-Image \cite{Clip} model; iii){\textbf{EffNet:}} A distance metric gathered from EfficientNet-B1 \cite{EffNet} model; iv){\textbf{SwAV:}} A distance metric gathered from SwAV-ResNet50 \cite{Swav} model.

Regarding the PixCorr and SSIM metrics, we resized the generated images from 512 × 512 resolution to match the 425 × 425 resolution of the ground-truth images in the NSD dataset. As for the remaining metrics, the preprocessing of generated images was performed based on the specific input characteristics of each network.

\begin{figure*}[!t]
\centering
\includegraphics[width=\linewidth]{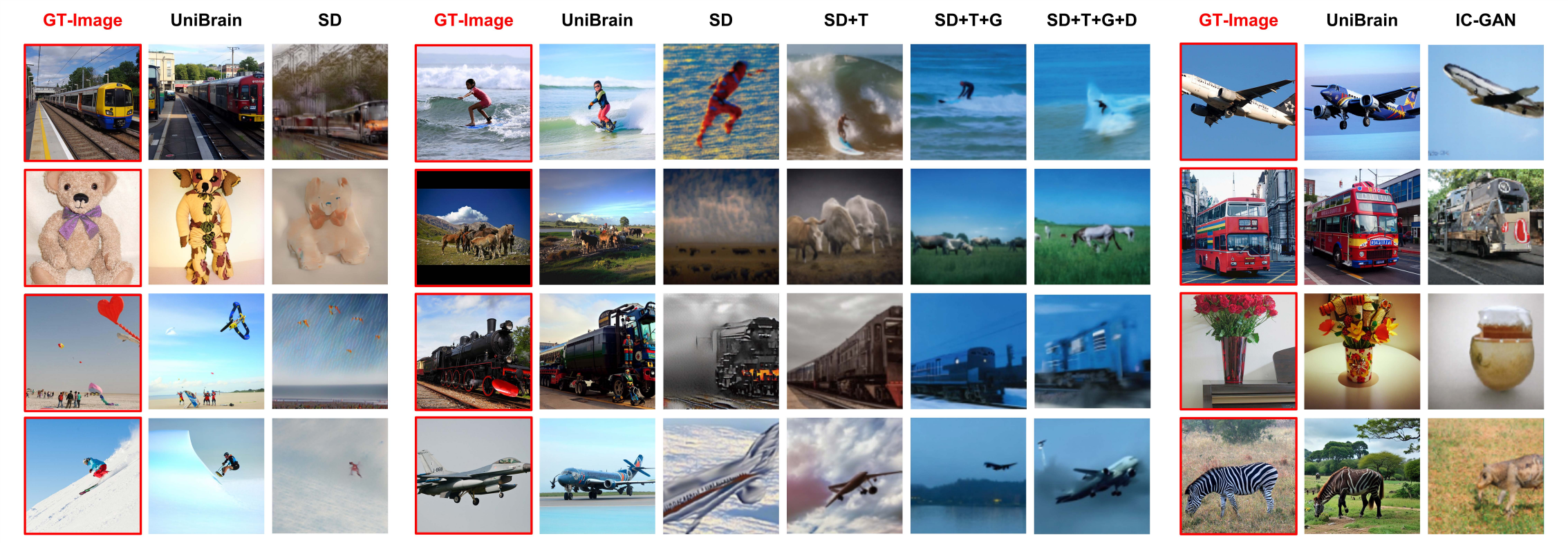}
\caption{Comparison of images reconstructed from UniBrain and current methods. since the presented test images in current methods are different, to be fair, comparisons between UniBrain and current models are separated. Left: Comparison between UniBrain ad SD; Middle: Comparison between UniBrain and SD-Based methods (SD, SD+T, SD+T+G, SD+T+G+D); Right: Comparison between UniBrain and IC-GAN. In each part, images in the red boxes are ground-true visual stimuli.} 
\label{fig_4}
\end{figure*}

\subsubsection{Text Metric}
\label{subsubsec: text_metric}

\textbf{Low-Level:} Low-level text features provide basic information about the structure and composition of the text but do not capture the deeper meaning or semantic relationships between words, measured by: i){\textbf{Meteor:}} The Meteor metric \cite{Meteor} provides a more robust evaluation by considering not only word overlap but also word order, synonymy, and other linguistic aspects that impact translation quality; ii){\textbf{Rouge:}} Rouge-1 and Rouge-L \cite{Rouge} are specific variants of the Rouge (Recall-Oriented Understudy for Gisting Evaluation) metric that focus on capturing the similarity between the generated and reference summaries.
    
\textbf{High-Level:} High-level text features are more complex and meaningful representations of text data that capture the context, relationships, and semantics of words and sentences, measured by: i){\textbf{CLIP:}} A two-way comparison of the output layer of the CLIP-Text \cite{Clip} model.   

A two-way comparison refers to the accuracy percentage in evaluations that determine whether the original image embedding is more akin to its corresponding brain embedding or to a randomly chosen brain embedding.
    
\section{Results}
\label{sec:results}

To the best of our knowledge, the proposed model UniBrain is the first one that unifies image reconstruction and captioning from human brain activity. Figure \ref{fig_3} presents some samples of generated images (Left) and captions (right) from fMRI. Overall, it is evident that the reconstructed images successfully preserve most of the layout and semantics present in the ground-truth images. However, disparities in pixel-level specifics still persist. In terms of image captioning, results reproduce the main compositions and semantics of the ground-truth captions and well describe the ground-truth images, while mismatched words still remain (marked with red crosses in Figure \ref{fig_3} (right)).

UniBrain is capable of generating images and captions with both low-level and high-level quality from visual-evoked fMRI, which gives us an opportunity to compare and analyze model results from different perspectives.
For instance, looking specifically at the last row in Figure \ref{fig_3}, `white cows' in the ground-truth caption matches the content in the ground-truth image, however, the `white' information is missed in the caption of Sub-1, which also matches the image of Sub-1 (cows are not white). In generated captions of Sub-5 and Sub-7, the whole phrase `white cows' are even replaced by the word `animals', which is semantically similar although it does not perfectly match the original phrase. Looking back into the generated images of Sub-5 and Sub-7, they do reproduce the basic layout and semantic information in a complex natural scene. In the Second row, the word `truck' is similar to `train' in both semantic and shape if we consider it a long truck or a truck train.
In the fourth row, reconstructed text results from Sub-2 and Sub-7 seem to be relatively dissimilar samples in semantics, though `surfboard' is similar to `skate' and `sled' in shape. In this case, we found that generated images were more consistent across subjects than generated captions.  The consistency of the generated results across subjects partly demonstrates the superior generalization performance of UniBrain, despite the fact that all training and test data came from a single subject.

\begin{table*}[htbp]
  \centering
  \caption{Compare Results of Image Reconstruction with Current Methods.}
    \renewcommand{\arraystretch}{1.3}
    \resizebox{\textwidth}{!}{
    \begin{tabular}{ccccc|cccc}
    \thickhline
    \multirow{2}[1]{*}{\textbf{Model}} & \multicolumn{4}{c}{\textbf{Low-Level}} & \multicolumn{4}{c}{\textbf{High-Level}} \\
    \cline{2-9} & \textbf{PixCorr $\uparrow$} & \textbf{SSIM $\uparrow$} & \textbf{AlexNet-2 $\uparrow$} & \textbf{AlexNet-5 $\uparrow$} & \textbf{Inception $\uparrow$} & \textbf{CLIP $\uparrow$} & \textbf{EffNet $\downarrow$} & \textbf{SwAV $\downarrow$} \\
    \hline
    Mind-Reader \cite{Mind-Reader} & -  & -  & -  & -  & 0.782 & -  & -   & - \\
    IC-GAN \cite{IC-GAN} & 0.150  & 0.325 & -  & -  & -  & - & 0.862 & 0.465 \\
    SD \cite{Takagi}   & - & - & 0.814 & 0.815 & 0.760 & 0.770 & - & - \\
    SD + T \cite{SD+TGD} & -  & -  & 0.852  & 0.914    & 0.858 & 0.841 & -  & - \\
    SD+T+G \cite{SD+TGD} & -  & -  & 0.880  & 0.929     & 0.864 & 0.853 & -  & - \\
    SD+T+G+D \cite{SD+TGD} & -  & -  & 0.858  & 0.921    & 0.843 & 0.866 & -  & - \\
    \hline
    UniBrain & \textbf{0.249} & \textbf{0.330} & \textbf{0.929} & \textbf{0.956} & \textbf{0.878} & \textbf{0.923} & \textbf{0.766} & \textbf{0.407} \\
    \thickhline
    \end{tabular}%
    }
  \label{tab_1}%
\end{table*}%

\begin{table*}[htbp]
  \centering
  \caption{Compare Results of Image Captioning with Current Methods.}
    \renewcommand{\arraystretch}{1.3}
    \begin{tabular}{cccccc}
    \thickhline
    \multirow{2}[1]{*}{\textbf{Subject}} & \multirow{2}[1]{*}{\textbf{Dataset}} & \multicolumn{3}{c}{\textbf{Low-Level}} & \textbf{High-Level} \\
    \cline{3-6}& &  \textbf{Meteor $\uparrow$} & \textbf{Rouge-1 $\uparrow$} & \textbf{Rouge-L $\uparrow$} &  \textbf{CLIP $\uparrow$}\\
    \hline
    Ridge-LSTM \cite{Ridge-LSTM} & GOD \cite{GOD} & -     & -     & -     & 46.4\% \\
    PT-LDM \cite{PT-LDM} & \cite{PT-LDM} & - & -     & 0.197 & - \\
    CNN-Transformer \cite{CNN-Transformer} & \cite{PT-LDM} & -     & -     & 0.201 & - \\
    \hline
    UniBrain & NSD \cite{NSD} & \textbf{0.169} & \textbf{0.245} & \textbf{0.222} & \textbf{85.3\%} \\
    \thickhline
    \end{tabular}%
  \label{tab_2}%
\end{table*}%

\subsection{Image Reconstruction Results}
\label{subsec: compare_vis}

We contrast the qualitative results of our model with current models in Figure \ref{fig_4}. We chose distinct images for each model to facilitate comparison, considering the variations in test images presented across their respective papers.

Being the inaugural research to leverage the NSD dataset for reconstruction purposes, Mind-Reader \cite{Mind-Reader} bears a resemblance to our approach in employing both text and image features as conditions. However, their methodology employed a StyleGAN2 model, as opposed to our utilization of LDMs. Note that the test set of Mind-Reader is different from ours, so we skip the qualitative comparison with Mind-Reader here and only make vague quantitative comparisons later.

Takagi \textit{et al.} \cite{Takagi} leverage a Stable Diffusion (SD) guided by text conditions to reconstruct images from the NSD dataset. While SD produces identifiable silhouettes, it falls short in terms of many qualitative aspects (i.e. low-level details, high-level semantics, and naturality) when compared to our model.
They recently published a new work \cite{SD+TGD} to discuss the improvement of the image generation effect after adding components of decoded text (T), GAN images (G), and depth images (D) on the basis of SD. These models are defined as SD+T, SD+T+G, and SD+T+G+D, respectively. However, as shown in Figure \ref{fig_4} (Middle), it seems that adding more components does not always lead to improved outcomes. There are still trade-offs that need to be made based on the generated performance. Nevertheless, these SD variant models are still lacking in low-level details and naturalness when compared to UniBrain.

The Instance-Conditioned GAN (IC-GAN) model \cite{IC-GAN} produces similar semantics but remains disparities in structural aspects such as shape and texture details, which makes the reconstruction results of IC-GAN less realistic when compared to UniBrain (Figure \ref{fig_4} (Right)). In addition, IC-GAN pretrained on ImageNet may be limited in complex natural scenes with multiple objects.

We compare the image reconstruction results of UniBrain with that of current models quantitatively via the visual metrics described in Section \ref{subsubsec: vis_metric}, including low-level and high-level metrics. Note that the metrics reported by previous models vary, but each model has at least one metric compared to ours. As shown in Table \ref{tab_1} ($\uparrow$ means higher is better, and $\downarrow$ means lower is better), UniBrain significantly outperforms all current models in all low-level and high-level metrics, thus achieving state-of-the-art performance.

\subsection{Image Captioning Results}
\label{subsec: compare_text}

Takada \textit{et al.} \cite{Ridge-LSTM} transform fMRI data from the GOD \cite{GOD} dataset into text features via Ridge regression and use the recurrent neural network with long short-term memory (LSTM) for caption generation. 
Huang \textit{et al.} \cite{PT-LDM} collect a visual-evoked fMRI dataset themselves and present the PT-LDM model to decode language from fMRI. Specifically, they use a convolutional neural network and a bidirectional GRU neural network to extract features from images and fMRI, which are merged in a GRU neural network to decode sentences. 
Based on the fMRI dataset collected by Huang \textit{et al.} \cite{PT-LDM}, Zhang \textit{et al.} \cite{CNN-Transformer} design a CNN-Transformer hybrid language decoding model to generate descriptive captions for visual stimuli. 

\begin{figure*}[!t]
\centering
\includegraphics[width=\linewidth]{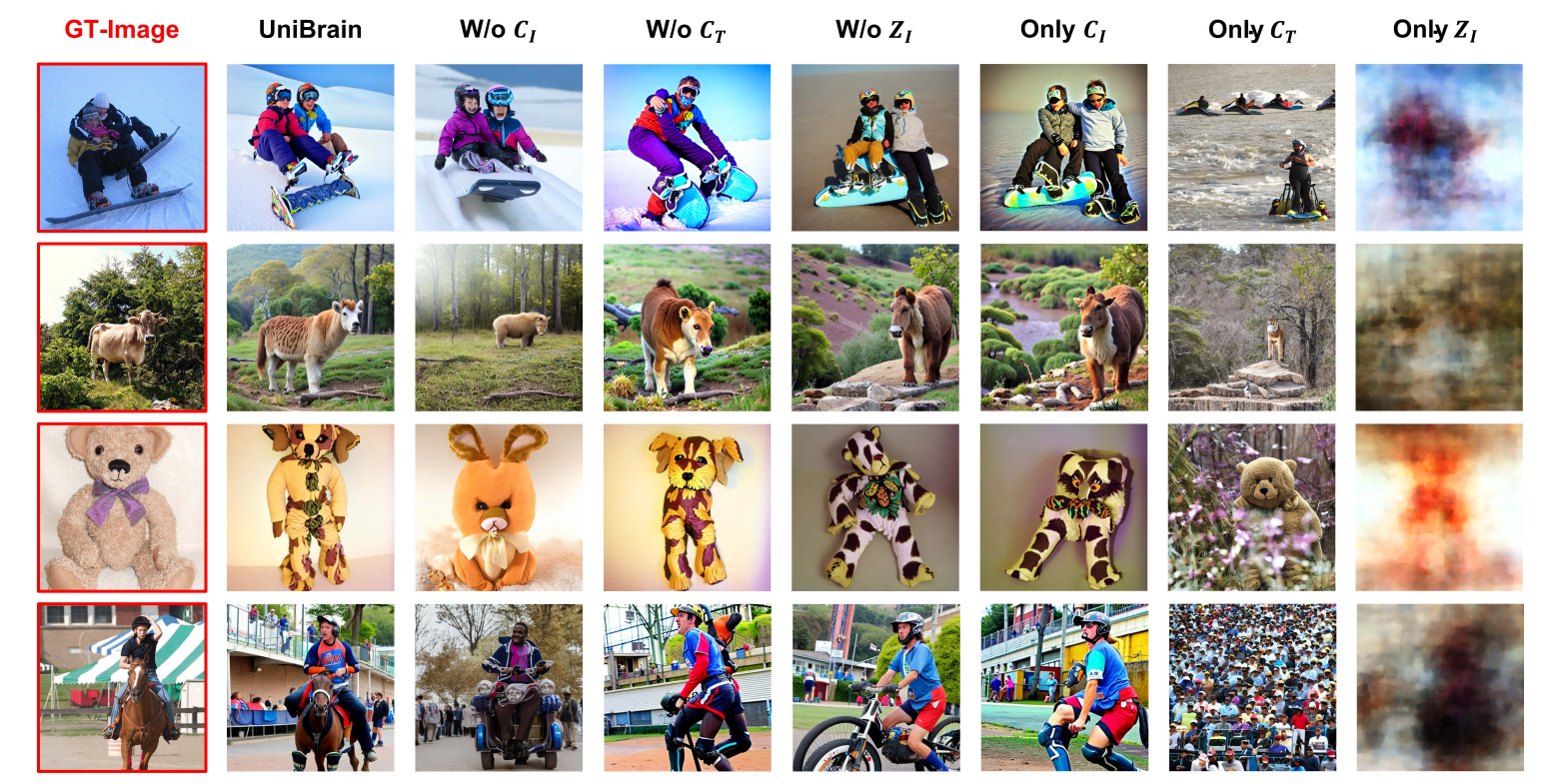}
\caption{Reconstructed images from UniBrain and its ablation models.`W/o $Z_I$' means removing the image latent representation  $Z_I$ from the initial UniBrain framework, similar to other 'W/o' ablation models. `Only $Z_I$' means only feed $Z_I$ to the diffusion process, similar to other 'Only' ablation models.}
\label{fig_5}
\end{figure*}

Note that UniBrain is the first model that performs image captioning tasks on the NSD dataset and utilizes LDMs for brain-image-text multimodal decoding. Therefore, we only make vague comparisons quantitatively between UniBrain and other previous models via text metrics described in Section \ref{subsubsec: text_metric}. As shown in Table \ref{tab_2}, we report the first image=  captioning result on the NSD dataset, which performs well on  both low-level and high-level metrics.

\subsection{Ablation Experiments}
\label{subsec: ablation}

In order to verify the effectiveness of each component in UniBrain, we conducted ablation experiments on Sub-1 for image reconstruction and captioning, respectively.
For image reconstruction, we present qualitative results in Figure \ref{fig_5} and report quantitative measures in Table \ref{tab_3}.

As shown in Figure \ref{fig_5}, Latent-Image $Z_I$ retains the rough outline and pixels of the original image (`Only $Z_I$' model). The `Only $C_I$' model shows that CLIP-Image features can preserve the global layout and basic semantic information of visual stimuli, such as `Two men on snowboards' (first row), and `A cow stands on the mountain' (second row). The `Only $C_T$' model fails in reconstructing images qualitatively, exhibiting that CLIP-Text features can not work alone but brings more high-level semantic details on top of CLIP-Image features (`Only $C_T$' vs `Only $C_I$' vs `W/o $Z_I$' ). UniBrain combines the advantages of each component and finally reconstructs images with both low-level details (pixels and contours) and high-level semantic information (object relationships, and contextual understanding) from fMRI.

\begin{table*}[htbp]
  \centering
  \caption{Ablation results of UniBrain for Sub-1 Image Reconstruction.}
    \renewcommand{\arraystretch}{1.3}
    \resizebox{\textwidth}{!}{
    \begin{tabular}{ccccc|cccc}
    \thickhline
    \multirow{2}[1]{*}{\textbf{Model}} & \multicolumn{4}{c}{\textbf{Low-Level}} & \multicolumn{4}{c}{\textbf{High-Level}} \\
    \cline{2-9} & \textbf{PixCorr $\uparrow$} & \textbf{SSIM $\uparrow$} & \textbf{AlexNet-2 $\uparrow$} & \textbf{AlexNet-5 $\uparrow$} & \textbf{Inception $\uparrow$} & \textbf{CLIP $\uparrow$} & \textbf{EffNet $\downarrow$} & \textbf{SwAV $\downarrow$} \\
    \hline
    Only $Z_I$ & \textbf{0.376} & \textbf{0.419} & 0.866 & 0.762 & 0.547 & 0.532 & 0.993 & 0.674 \\
    Only $C_T$ & 0.015 & 0.209 & 0.633 & 0.788 & 0.761 & 0.837 & 0.872 & 0.598 \\
    Only $C_I$ & 0.089 & 0.289 & 0.830 & 0.926 & 0.863 & 0.915 & 0.800 & 0.436 \\
    W/o $Z_I$ & 0.062 & 0.289 & 0.833 & 0.927 & 0.864 & 0.927 & 0.778 & 0.424 \\
    W/o $C_T$ & 0.270 & 0.324 & 0.943 & 0.968 & 0.875 & 0.914 & 0.787 & 0.420 \\
    W/o $C_I$ & 0.331 & 0.363 & 0.913 & 0.926 & 0.849 & 0.869 & 0.811 & 0.460 \\
    \hline
    \textbf{UniBrain} & 0.299 & 0.342 & \textbf{0.954} & \textbf{0.969} & \textbf{0.890} & \textbf{0.930} & \textbf{0.757} & \textbf{0.396} \\
    \thickhline
    \end{tabular}%
    }
  \label{tab_3}%
\end{table*}%

\begin{table}[htbp]
  \centering
  \caption{Ablation results of UniBrain for Sub-1 Image Captioning.}
    \renewcommand{\arraystretch}{1.3}
    \scalebox{0.87}{
    \begin{tabular}{ccccc}
    \thickhline
    \multirow{2}[1]{*}{\textbf{Subject}} & \multicolumn{3}{c}{\textbf{Low-Level}} & \textbf{High-Level} \\
    \cline{2-5} & \textbf{Meteor$\uparrow$} & \textbf{Rouge-1$\uparrow$} & \textbf{Rouge-L$\uparrow$} &  \textbf{CLIP$\uparrow$}\\
    \hline
    Only $Z_T$ & \textbf{0.185} & \textbf{0.262} & \textbf{0.233} & 77.2\% \\
    Only $C_T$ & 0.118 & 0.218 & 0.208 & 67.7\% \\
    Only $C_I$ & 0.128 & 0.178 & 0.157 & 82.8\% \\
    W/o $Z_T$ & 0.161 & 0.234 & 0.209 & 84.9\% \\
   W/o $C_T$ & 0.136 & 0.192 & 0.169 & 83.4\% \\
    W/o $C_I$ & 0.113 & 0.214 & 0.205 & 68.0\% \\
    \hline
    \textbf{UniBrain} & 0.170 & 0.247 & 0.225 & \textbf{86.1\%} \\
    \thickhline
    \end{tabular}%
    }
  \label{tab_4}%
\end{table}%

Table \ref{tab_3} further validates our conclusions from a quantitative perspective. Specifically, the `Only $Z_I$' model achieves the best performance in pixel indicators, and the removal of latent information (`W/o $Z_I$') makes UniBrain drop sharply in all Low-Level indicators, which proves that $Z_I$ retains most of the low-level features. Furthermore, we noticed that: i)`Only $C_I$' significantly outperforms the `Only $C_T$' model in all low-level and high-level metrics; ii)`W/o $C_I$' performs worse than `W/o $C_T$' in all high-level metrics; iii)`W/ o $C_I$' is even worse than `Only $C_I$' in all high-level metrics; iv) `W/o $Z_I$' outperforms `Only $C_I$' in all high-level metrics. These results reveal that CLIP-Image retains more basic information about the original image during image reconstruction, and the semantic features extracted by CLIP-Text need to be built on top of this. That is, CLIP-Image pays more attention to the overall semantic information, while CLIP-Text pays more attention to the detailed semantic information. We all know that details need to work better on the whole. Interestingly, though $Z_I$ performs poorly on high-levels matrics, it boost the performance of $C_T$ (`Only $C_T$' vs `W/o $C_I$') and $C_I$ (`Only $C_I$' vs `W/o $C_T$').

Note that, even when combined with $C_T$ and $Z_I$, as with Takagi \textit{et al.} \cite{Takagi}, all quantitative high-level performance of the `W/o $C_I$' model is still worse than the `Only $C_I$' model. This phenomenon explains why UniBrain is superior to SD \cite{Takagi}, since SD ignores the $C_I$ features. The direct comparison between UniBrain and the `w/o $C_I$' model more intuitively demonstrates the significant improvement that $C_I$ brings to UniBrain. UniBrain achieves the best in all indicators except pixel indicators (PixCorr and SSIM, UniBrain is the second best), indicating the effectiveness of fusing different components in UniBrain.

For image captioning, we report quantitative measures for UniBrain and its ablation models in Table \ref{tab_4}. Text is made up of words, unlike images made up of pixels. Words are inherently semantic, pixels are not. Therefore, text latent has high semantic features. It is worth noting that the high-level metric of the `Only $Z_T$' model scores even better than that of the `Only $C_T$' model. Interestingly, as in image reconstruction, $C_I$ are more important than $C_T$ features in image captioning, and the gap is more pronounced. This also explains why the mixing ratio $mix=0.9$ is the best-performed value for image captioning. But even so, the performance of UniBrain has been significantly improved compared to the `W/o $C_T$' model. This phenomenon is also the same as that found in image reconstruction. $C_T$ works poorly when it works alone, but it can be used as the performance propellant of $C_I$ (`Only $C_T$' vs `Only $C_I$' vs `W/o $Z_T$') . The caption generation performance of UniBrain degrades when any component is missing. Except for the optimal performance of the `Only $Z_T$' model in low-level indicators, UniBrain achieved the best performance in all indicators compared with other models.

\begin{figure}[!t]
\centering
\includegraphics[width=\linewidth]{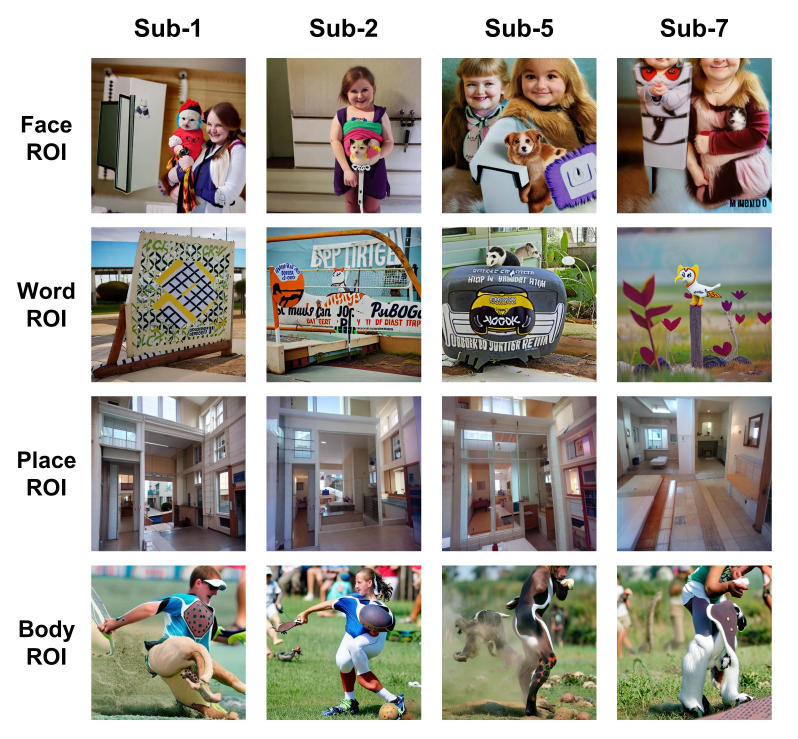}
\caption{Images reconstructed from synthetic fMRI voxels derived by activating specific regions-of-interests (ROIs).  The rows present ROIs that are specified with functional localization experiments (Face-ROI, Word-ROI, Place-ROI, Body-ROI).}
\label{fig_6}
\end{figure}

\subsection{ROI Analysis}
\label{subsec: roi}

Beyond brain decoding, UniBrain has the potential to shed light on the functional characteristics of particular regions-of-interest (ROIs) within the human brain. The interpretation of ROI-optimal images for high-level ROIs defined by function is straightforward, as they typically align with the known category preference of each region. As shown in Figure \ref{fig_6}, UniBrain produced various face images from Face-ROI, involving both human and animal faces. Pseudo-words and characters on signs or objects were produced within the Word-ROI except for Sub-7. Regarding the Place-ROI, UniBrain generated realistic interior scene architectural layouts. In the end, UniBrain reconstructed body parts of humans or animals that are participating in active movements from Body-ROI. Directly visualizing the "optimal" stimulation of a given functional brain region can improve our understanding of how human brain responds to external stimuli. 

\section{Conclusion}

In this work, we propose a multi-task multi-modality model termed UniBrain for image reconstruction and captioning from visual-evoked brain activity measured by fMRI. We delved deeper into the capabilities of LDM in integrating brain, image, and text decoding,  facilitating the accomplishment of image reconstruction and captioning tasks within a unified model. This effort allows us to fully analyze how human brain processes and understands external visual stimuli. We compared UniBrain with current models for visual-evoked fMRI-based image reconstruction and captioning respectively. Empirical evaluation demonstrated that UniBrain outperforms other models qualitatively and quantitatively in each task. We made a detailed analysis by combining and contrasting the generated results of images and captions. In addition, we conducted ablation experiments and functional ROI analysis to further exhibit the superiority of UniBrain and provide comprehensive insight for visual-evoked brain decoding. A unified model of multi-task and multi-modality will be closer to the way the brain understands the outside world. We anticipate that this work will serve as a wellspring of inspiration for future investigations.

{\small
\bibliographystyle{ieee_fullname}
\bibliography{egbib}
}

\end{document}